\documentclass[runningheads]{llncs}
\usepackage{graphicx}
\usepackage{comment}
\usepackage{multirow}
\usepackage{algorithmic}
\usepackage{algorithm}
\usepackage{capt-of}
\usepackage{amsmath,amssymb} 
\usepackage{color}
\usepackage{wrapfig}
\usepackage{subfigure}

\usepackage{xspace}
\makeatletter
\DeclareRobustCommand\onedot{\futurelet\@let@token\@onedot}
\def\@onedot{\ifx\@let@token.\else.\null\fi\xspace}

\def\etal{{et al}\onedot}
\makeatother

\newsavebox\tmpbox
\newcommand{\tabincell}[2]{\begin{tabular}{@{}#1@{}}#2\end{tabular}} 
\newcommand{\blocka}[2]{\multirow{2}{*}{\(\left[\begin{array}{c}\text{3$\times$3, #1}\\[-.1em] \text{3$\times$3, #1} \end{array}\right]\)$\times$#2}
}
\begin{document}
\pagestyle{headings}
\mainmatter

\title{Knowledge distillation via adaptive instance normalization}
\titlerunning{Knowledge distillation via adaptive instance normalization}
\author{Jing Yang \inst{1} \and
       Brais Martinez\inst{2} \and
       Adrian Bulat\inst{2} \and
       Georgios Tzimiropoulos\inst{1,3}}
\authorrunning{Yang et al.}
\institute{Computer Vision Laboratory,
University of Nottingham, UK \and 
Samsung AI Center, Cambridge, UK \and
Queen Mary University of London, London, UK\\
\email{jing.yang2@nottingham.ac.uk,
brais.mart@gmail.com,
adrian@adrianbulat.com,  georgios.t@samsung.com/g.tzimiropoulos@qmul.ac.uk}}

\maketitle

\begin{abstract}
This paper addresses the problem of model compression via knowledge distillation. To this end, we propose a new knowledge distillation method based on transferring feature statistics, specifically the channel-wise mean and variance, from the teacher to the student. Our method goes beyond the standard way of enforcing the mean and variance of the student to be similar to those of the teacher through an $L_2$ loss, which we found it to be of limited effectiveness. Specifically, we propose a new loss based on adaptive instance normalization to effectively transfer the feature statistics. The main idea is to transfer the learned statistics back to the teacher via adaptive instance normalization (conditioned on the student) and let the teacher network ``evaluate'' via a loss whether the statistics learned by the student are reliably transferred. We show that our distillation method outperforms other state-of-the-art distillation methods over a large set of experimental settings including different (a) network architectures, (b) teacher-student capacities, (c) datasets, and (d) domains.
\end{abstract}

\section{Introduction}
\label{section:Introduction}
Recently, there has been a great amount of research effort to make Convolutional Neural Networks (CNNs) lightweight so that they can be deployed in devices with limited resources. To this end, several approaches for model compression have been proposed including network pruning~\cite{han2015deep,lebedev2016fast}, network quantization~\cite{rastegari2016xnor,wu2016quantized}, knowledge transfer/distillation~\cite{hinton2015distilling,zagoruyko2016paying}, and neural architecture search~\cite{zoph2016neural,liu2018darts}. Knowledge distillation, one of the most popular methods for model compression, refers to transferring knowledge
from one network (the so-called ``teacher'') to another
(the so-called ``student''). Typically, the teacher is a high capacity model capable of achieving high accuracy, while the student is a compact model with much fewer parameters, thus also requiring much less computation. The goal of knowledge
distillation is to use the teacher to train a compact student
model with similar accuracy to that of the teacher. This paper proposes a surprisingly simple and effective method for knowledge distillation via borrowing ideas from adaptive instance normalization~\cite{huang2017arbitrary}.

\begin{figure}[t]
\centering
\includegraphics[height=2.4in,trim={0cm 0.cm 0cm 0.cm},clip]{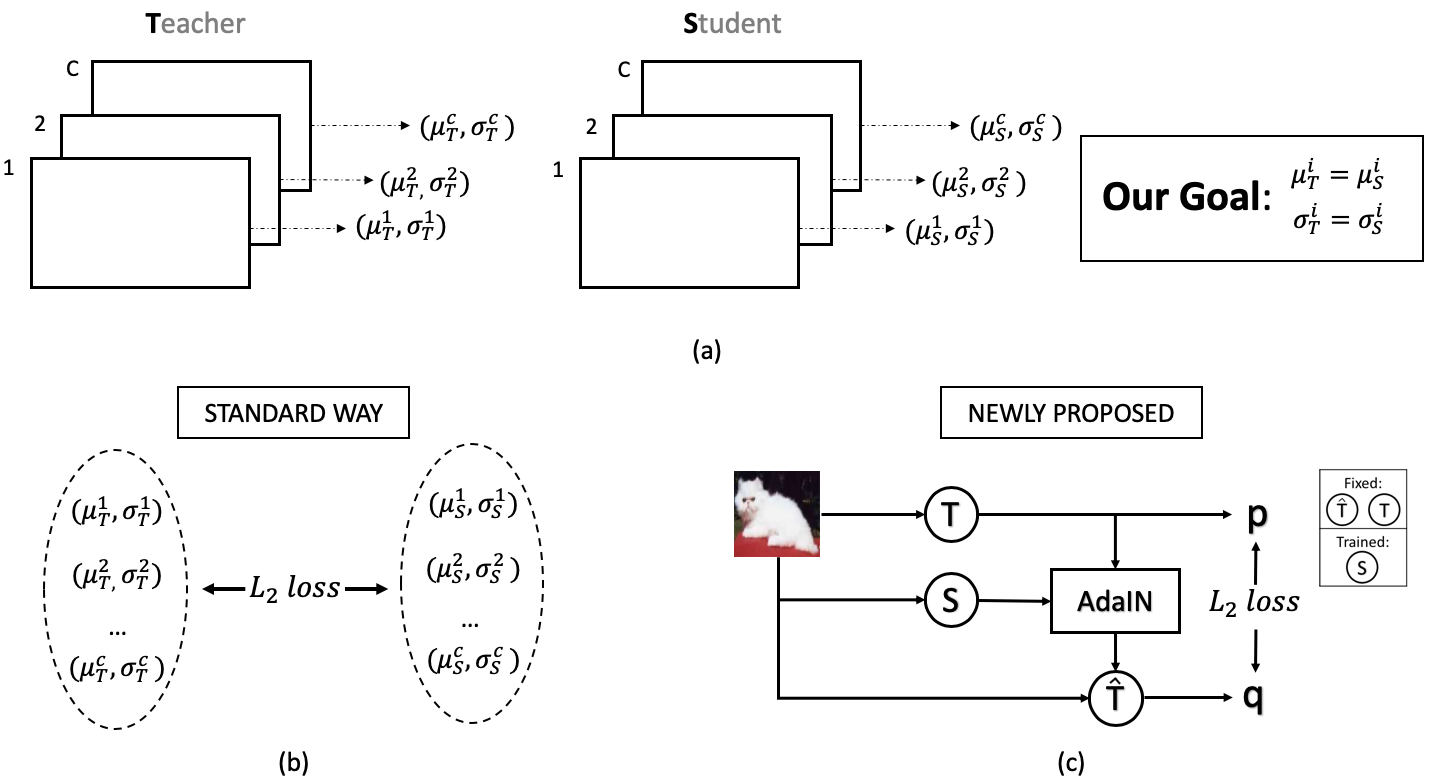}
\caption{Our method performs knowledge distillation by aligning the channel-wise statistics mean and variance between the feature maps extracted from the teacher and the student. In (b), we show the standard direct way to achieve this through an $L2$ loss. In (c), we propose a new mechanism: \textbf{transfer the learned by the student statistics back to the teacher via adaptive instance normalization} and let the teacher network ``evaluate'' via an $L2$ loss, whether the statistics learned by the student are reliably transferred. Note that only the student is trained in (c). $\hat{T}$ here denotes the teacher after the copying the statistics from the student and shares the same weights as the original teacher.}
\label{fig:insight}
\vspace{-6mm}
\end{figure}

The rationale behind knowledge distillation can be explained from an optimization perspective: there is evidence that high capacity models (i.e. the teachers) can find good local minima due to over-parameterization~\cite{du2018power,soltanolkotabi2018theoretical}. In knowledge distillation, such models are used to facilitate the optimization of lower capacity models (i.e. the students) during training. For example, in~\cite{hinton2015distilling}, the soft outputs of the teacher provide extra supervisory signals of inter-class similarities which facilitate the learning of the student. In a similar fashion, intermediate representations extracted from the teacher such as feature tensors~\cite{romero2014fitnets} or attention maps~\cite{zagoruyko2016paying} have been used to define loss functions used to facilitate the optimization of the student. 

This paper proposes a surprisingly unexplored idea for knowledge distillation: for a given training example, transfer feature statistics, in particular, the channel-wise mean and variance from the teacher to the student. The usefulness of this kind of statistics transfer has been previously demonstrated in learning deep generative models~\cite{li2015generative}, domain adaptation~\cite{li2016revisiting}, as well as style transfer \cite{huang2017arbitrary,li2017demystifying} but to our knowledge, it is the first time that they are used for knowledge distillation. 
Equally importantly, our method goes beyond the standard way of enforcing the mean and variance of the student to be similar to those of the teacher through an $L_2$ loss as shown in Figure~\ref{fig:insight}(b). This is motivated by the fact that, in practice, we found that an $L_2$ loss is not sufficient for the statistics to be effectively transferred. To alleviate this, we propose a new supervision loss for statistics transfer, inspired by the adaptive instance normalization of~\cite{huang2017arbitrary}: the main idea is to \textbf{transfer the learned statistics back to the teacher} via adaptive instance normalization (conditioned on the student) and let the teacher ``evaluate'' via a loss whether the statistics learned by the student are reliably transferred (this is denoted as $L_{2}$ in Figure~\ref{fig:insight}(c)). More specifically, we use instance normalization to normalize the teacher's feature tensors to zero mean and unit variance, and then scale and shift them using the statistics provided by the student. Finally, a new loss is applied between the output of this network and that of the original teacher, to make this newly affine-transformed feature tensor able to obtain similar performance to that of the original teacher's feature tensor. This loss is used to enforce that the statistics of the student do not change the predictions of the teacher, providing an additional supervision signal that enables effective feature statistics transfer from the teacher to the student. In summary, \textbf{our contributions} are:  
\begin{enumerate}
    \item 
    We propose a new knowledge distillation method based on transferring feature statistics, specifically the channel-wise mean and variance, from the teacher to the student.
    \item We propose a new mechanism to effectively transfer the feature statistics which uses adaptive instance normalization to transfer the learned statistics back to the teacher and then applies a loss on the output of the teacher network which further enforces that the statistics are reliably transferred. 
    \item We show that our method consistently outperforms other state-of-the-art distillation methods 
    over a large set of experimental settings including different (a) network architectures (Wide ResNets, ResNets, MobileNet, MobileNetV2), (b) teacher-student capacities, (c) datasets (CIFAR-10/100, ImageNet), and (d) domains (real-to-binary).
\end{enumerate}

\section{Related Work}
\label{section:Related Work}
Knowledge distillation with neural networks, pioneered by Hinton \etal \cite{hinton2015distilling} and Buciluă \etal \cite{buci2006model}, is a transfer learning method that enhances the accuracy of a lightweight student network based 
on the ``knowledge'' provided by a powerful teacher network.
In \cite{hinton2015distilling}, knowledge is defined as the teacher's soft outputs after the final softmax layer.
The soft outputs carry richer information than one-hot labels because they provide extra supervision signals in terms of the inter-class similarities learned by the teacher. This idea is further explored by other works. 
Zhou \etal \cite{zhou2018rocket} proposed a novel method which introduces a booster net which is trained together with the student network and a logit matching loss for supervising the student's prediction. Another work that also trains the student and teacher simultaneously is \cite{zhang2018deep}, in which each student is trained with a conventional loss for supervised learning, and with a mimicry loss that aligns each student’s class posterior with the class probabilities of a teacher.
Recently, \cite{Cho_2019_ICCV} found that very accurate networks are ``too good'' to be good teachers and proposed to mitigate this with early stopping of the teacher's training.

In a similar fashion to~\cite{hinton2015distilling}, intermediate representations extracted from the teacher such as feature tensors~\cite{romero2014fitnets} or attention maps~\cite{zagoruyko2016paying} have been used to define loss functions used to facilitate the optimization of the student. Trying to match the whole feature tensor, as in FitNets~\cite{romero2014fitnets}, is hard and, in certain circumstances, such an approach may adversely affect the performance and convergence of the student. To relax the assumption of FitNet, Attention Transfer (AT) was proposed in~\cite{zagoruyko2016paying} where knowledge takes the form of attention maps which are summaries of the energies of the feature tensors over the channel dimension. 
An extension of~\cite{zagoruyko2016paying} using Maximum Mean Discrepancy of the network activations as a loss term for distillation was proposed in ~\cite{huang2017like}. 

In \cite{heo2019knowledge}, the authors advocate the use of the activation boundary of neurons (rather than their exact output values), and propose a knowledge transfer method via distillation of activation boundaries formed by hidden neurons. This work is further improved by ~\cite{heo2019comprehensive}, which studies the location within the network at which feature distillation should be applied and proposes margin ReLU and a specifically designed distance function that transfers only the useful (positive) information from the teacher to the student. 

Another line of knowledge distillation methods focus on exploring transferring the relationship between features, rather than the actual features themselves.
In~\cite{yim2017gift}, feature correlations are captured by computing the Gram matrix of features across layers for both teacher and student and then applying an $L_2$ loss on pairs of teacher-student Gram matrices. The limitation of this work is the high computational cost, which is addressed to some extent by~\cite{lee2018self} by compressing the feature maps by singular value decomposition. The authors of~\cite{park2019relational} propose a relational knowledge distillation method which computes distance-wise and angle-wise relations of each embedded feature vector. This idea is further explored in~\cite{peng2019correlation} and~\cite{liu2019knowledge}. In \cite{peng2019correlation}, Taylor series expansion is proposed to better capture the correlation between multiple instances. In \cite{liu2019knowledge}, the instance feature and relationships are considered as vertexes and edges respectively in a graph and instance relationship graph is proposed to model the feature space transformation across layers. Inspired by the observation that semantically similar inputs should have similar activation patterns, \cite{tung2019similarity} proposes a similarity-preserving knowledge distillation method which guides the student to mimic the teacher with respect to generating similar or dissimilar activations. More recently, \cite{tian2019contrastive} formulates 
structural knowledge extraction as as contrastive learning and proposes to train a student to capture correlations and higher order output dependencies in representation of the data.

From the above review it can be deduced that there is no method for knowledge distillation based on transferring feature statistics (channel-wise mean and variance) and adaptive instance normalization as proposed in our work. 

\vspace{-4mm}
\section{Method}
This section introduces our method as follows: Section \ref{ssec:overview} provides an overview of our method. Section \ref{ssec:Statistic Matching} describes the proposed feature statistics transfer for knowledge distillation using a standard $L_2$ loss. Section \ref{ssec:AIN} describes the newly proposed mechanism for statistics transfer via adaptive instance normalization. Section \ref{ssec:alg} provides the training procedure of our method.

\begin{figure}[t]
\centering
\includegraphics[height=2.3in,trim={0cm 0.cm 0cm 0.cm},clip]{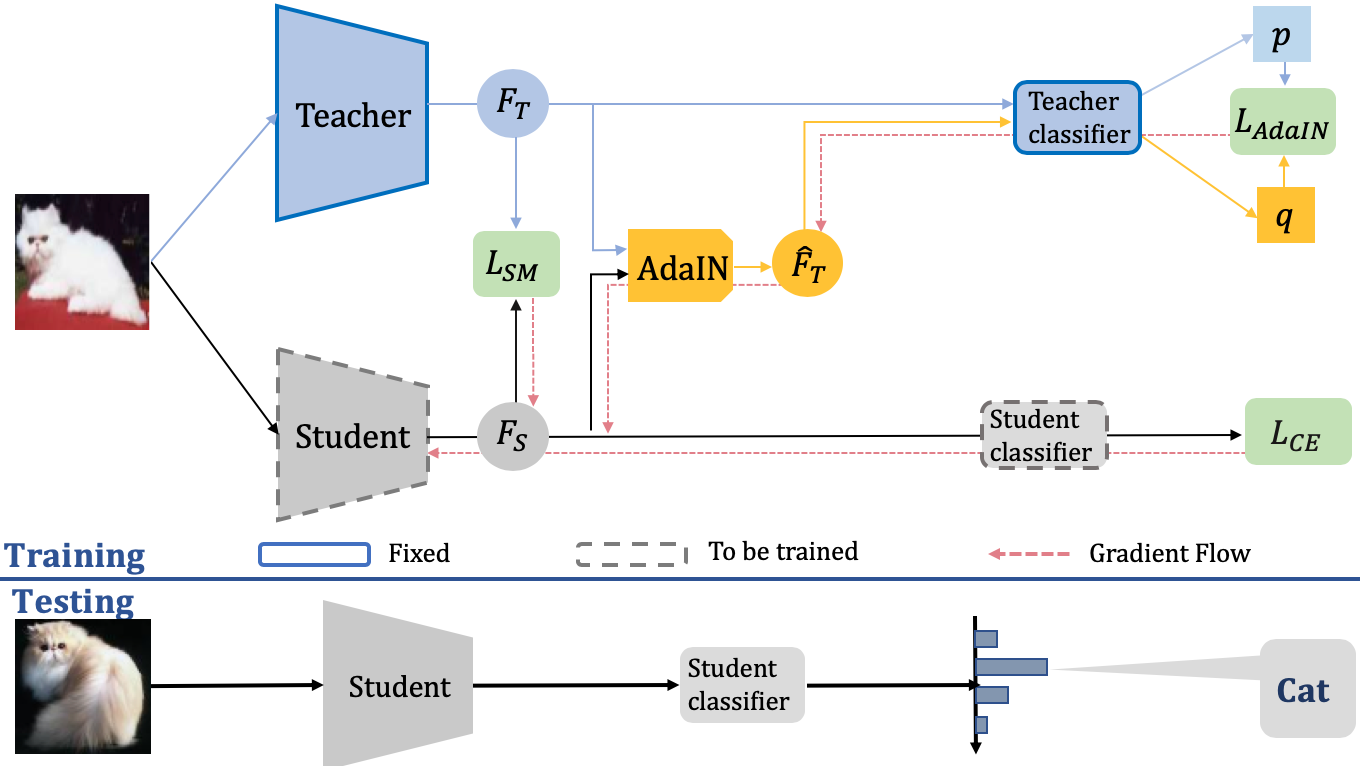}
\caption{Overview of our method. For simplicity, it is shown for only one layer of the teacher and student networks. $L_{SM}$: an $L_2$ loss between the channel-wise mean and variance of the teacher's and student's feature maps. $L_{AdaIN}$: AdaIN is applied to the teacher using the student's statistics, to obtain the affine-transformed feature $\hat{F}_{T}$. An $L_2$ loss is applied between the output of this network $q$ and that of the original teacher $p$, enforcing that the student's statistics do not change the teacher's predictions. $L_{CE}$: loss based on ground truth labels.} 
\vspace*{-10pt}
\label{fig:overview}
\end{figure}

\subsection{Overview}
\label{ssec:overview}
Figure~\ref{fig:overview} shows an overview of our method. The teacher network is pretrained and fixed during training the student. We train our student using three losses:
\begin{equation}
L = L_{CE} + \alpha L_{SM} + \beta L_{AdaIN}, \label{eq:allloss} 
\end{equation}
where $\alpha$ and $\beta$ are the weights used to scale the losses. 
$L_{CE}$ is the standard loss based on ground truth labels for the task in hand (e.g. cross-entropy loss for image classification). $L_{SM}$ enables feature statistics transfer using a standard $L_2$ loss (see Sec.~\ref{ssec:Statistic Matching}). This is simply the $L_2$ distance between the channel-wise mean and variance of feature maps extracted from the teacher and the student for various layers of the networks. 
Although effective to some extent, the $L_2$ is detached from the final task (e.g. classification), and the measure of goodness is an arbitrary isotropic distance. In order to capture the complex relations between the different statistics and the final performance of the network, we propose a novel loss inspired by adaptive instance normalization, termed $L_{AdaIN}$. This loss is used to enforce that the statistics of the student do not change the predictions of the teacher. The newly proposed loss based on adaptive instance normalization is computed as follows: a training example is fed to both the student and the teacher. Denote the output of the teacher as $p$. The same input is again fed to the teacher, but now for any pair of corresponding layers of the teacher and the student, we apply AdaIN to the teacher using the student's statistics. This yields a prediction $q$. $L_{AdaIN}$ is a loss applied to $p$ and $q$ enforcing them to be similar. $L_{SM}$ and $L_{AdaIN}$ are described in detail in the following subsections. 
\vspace{-4mm}
\subsection{Statistics Transfer via Statistics Matching}
\label{ssec:Statistic Matching}
Given an input image $\mathbf{x}$, denote the feature tensor produced by the teacher network $T$ at $i$-th layer as $F^{i}_{T} \in R^{C_i \times H_i \times W_i}$, where $C$ is the number of output channels, and $H$ and $W$ are spatial dimensions. Similarly, we denote by $F^{j}_{S}\in R^{C_j \times H_j \times W_j}$ the feature tensor at $j$-th layer produced by the student from the same input. As typically done in feature matching literature~\cite{zagoruyko2016paying,kim2018paraphrasing}, we choose layers $i$ and $j$ to be from the same spatial resolution, i.e. $H_i = H_j=H$ and $ W_i = W_j=W$. Moreover, for simplicity, we assume $C_i=C_j=C$, noting that when this is not the case we use a $1\times1$ convolution to make the two feature maps have the same number of channels. For simplicity, let us drop the dependence of $F^{i}_{T}$ and $F^{j}_{S}$ on $i$ and $j$. The channel-wise mean and variance for $F_{T}$ are:
\begin{equation}
\mu_T^{c} =\frac{1}{HW} \sum_{h=1}^{H}\sum_{w=1}^{W}F^{chw}_{T},
\end{equation}

\begin{equation}
\sigma_T^{c} =\sqrt{\frac{1}{HW}\sum_{h=1}^{H}\sum_{w=1}^{W}(F^{chw}_{T}-\mu_T^{c})^2+\epsilon},
\end{equation}
while, similarly, we can calculate $\mu_S^{c}$ and $\sigma_S^{c}$ from $F_S$. 
To match the statistics of $F_S$ with those of $F_T$ we can use the following $L_2$ loss:
\begin{equation}
L_{SM}(F_{T},F_{S})=\frac{1}{C}\sum _{c=1}^{C}\left( (\mu_T^{c}-\mu_S^{c})^2+(\sigma_T^{c}-\sigma_S^{c})^2 \right)
\end{equation}
Finally, feature statistics transfer based on statistics matching is obtained by summing over all $(F_{T},F_{S})$ pairs:
\begin{equation}
L_{SM} = \sum_{(F_{T},F_{S})}L_{SM}(F_{T},F_{S}). \label{eq:stloss}
\end{equation}

We found $L_{SM}$ to be effective not only in terms of statistics transfer but also in terms of improving the accuracy of the student network (see Table~\ref{table:ablation study}). 
\vspace{-4mm}
\subsection{Statistics Transfer via Adaptive Instance Normalization}
\label{ssec:AIN}
One disadvantage of $L_{SM}$ introduced in the previous section and, in general, of all feature matching losses e.g.~\cite{romero2014fitnets,zagoruyko2016paying}, is that it treats each channel dimensions in the feature space independently, and ignores the structural dependencies across channels for the final classification. This is in contrast to the original logit matching loss proposed by Hinton \etal in~\cite{hinton2015distilling} which directly targets classification accuracy. In this section, we tackle this problem by introducing a new mechanism based on adaptive instance normalization that uses the transferred statistics to directly minimize an auxiliary classification loss.

Adaptive instance normalization (AdaIN) was firstly proposed for arbitrary style transfer~\cite{huang2017arbitrary}, and since then it has been popular for several vision tasks, including unsupervised domain adaptation~\cite{li2016revisiting}, few-shot learning of head models~\cite{zakharov2019few} and image segmentation~\cite{sofiiuk2019adaptis}. Given a content (image) input and style input, AdaIN normalizes the content to the input style by simply aligning the channel-wise mean and variance of the content input to match those of the style.

In our case, $F_{T}$ functions as content, which is conditioned on $F_{S}$, serving as the input style. Specifically, instance normalization $\frac{F_{T}-\mu_T}{\sigma_T}$ normalizes each channel to be zero-mean and of unit variance. Then, the affine parameters $\mu_S$ and $\sigma_S$ provided by student network are used to map the normalized feature to the student feature space. Overall, by applying adaptive instance norm to $F_{T}$, we obtain the new normalized representation $\hat{F}_{T}$ as: 
\begin{equation}
\hat{F}_{T} =\sigma_S \frac{F_{T}-\mu_T}{\sigma_T}+\mu_S.
\label{eq:AIN}
\end{equation}
Let us denote by $p$ the output of the original teacher network when fed with some input image and by $q$ the output of the teacher network for the same input after applying adaptive instance normalization. Then, we can apply an $L_2$ loss between $p$ and $q$ so that $\sigma_S$ and $\mu_S$ from $\hat{F}_{T}$ are trained via back-propagation:
\begin{equation}
L_{AdaIN}(p,q) = \left \| p-q \right \|^2. \label{eq:LAdaIN}
\end{equation}
Notably, all other parameters of the teacher network remain frozen. Also note that if $p = q$, i.e. if the teacher network after adaptive instance norm is able to achieve the same performance as the original one, then this implies that $\hat{F}_{T} = F_{T}$. In this case, it is trivial to show that $\mu_S=\mu_T$, and $\sigma_S=\sigma_T$ which ensures the desirable statistics transfer from the teacher to the student. Finally, losses other than $L_2$ for $L_{AdaIN}$ are explored in the supplementary material.
\vspace{-4mm}
\subsection{Training algorithm}
\label{ssec:alg}
Overall, the training procedure of our method is provided in Algorithm \ref{alg}.
\vspace{-4mm}

\begin{algorithm}[!htb]
\begin{algorithmic}
\STATE \textbf{Input:} Teacher network $T$, Student network $S$, input image $\mathbf{x}$, ground truth label $y$.
\vspace{-3mm}
\begin{enumerate} \setlength{\itemsep}{-\itemsep}
\item Input $\mathbf{x}$ to $S$ to obtain prediction $\hat{y}$, cross entropy task loss $L_{CE}=\mathcal{H}(\hat{y},y)$ and $F_{S}$;
\item Input $\mathbf{x}$ to $T$ to obtain prediction $p$ and $F_{T}$; Compute statistics matching loss $L_{SM}$ from Eq.~(\ref{eq:stloss}).
\item Use Eq.~\ref{eq:AIN} (adaptive instance norm) to obtain the new feature $\hat{F}_{T}$. Input $\hat{F}_{T}$ to the teacher network to obtain $q$ and finally compute the loss $L_{AdaIN}$ from Eq.~(\ref{eq:LAdaIN}).
\item Update $S$ by optimizing Eq.~(\ref{eq:allloss}) 
\end{enumerate}
\vspace{-3mm}
\STATE \textbf{Output:} the updated $S$.
\end{algorithmic}
\caption{Knowledge distillation via AdaIN}
\label{alg}
\end{algorithm}

\section{Ablation Studies}
\label{sectoin:as}
In this section, we study and explore the influence of certain design choices on the overall accuracy of the proposed method deriving in the process the optimal configuration used for the rest of our experiments. In particular, Section~\ref{ssec:losses} offers an analysis of the effectiveness of each loss on the student's accuracy while in Section~\ref{ssec:position} we study the optimal placement of the statistics matching loss inside the network. As common in literature~\cite{kim2018paraphrasing}, we performed our ablation studies on CIFAR-100 (see also Section~\ref{ssec:CIFAR100}) using a Wide ResNet (WRN) for both the student and the teacher (the architecture is described in Table~\ref{table:WRNBlock}). Here, our teacher is WRN-40-4 and student is WRN-16-4.

\begin{table}[t]
\caption{Structure of the Wide ResNet (WRN) networks used in our experiments. $c$ denotes number of classes. For CIFAR-10, $c=10$. For CIFAR-100, $c=100$. $d=(D-4)/6$.}
\begin{center}
\vspace{-2mm}
\begin{tabular}{c|c|c|c}
\hline
\tabincell{c}{Group Name} & \tabincell{c}{Output size} & WRN-$D$-$k$   &WRN-16-4 \\ \hline
conv1& 32x32 &\multicolumn{2}{c}{3x3,16}  \\ \hline
\multirow{3}{*}{conv2}
& \multirow{3}{*}{32x32}
&\blocka{16$k$}{$d$}
&\blocka{64}{2}\\
  &   & \\
\hline
\multirow{3}{*}{conv3}& 
\multirow{3}{*}{16x16} 
& \blocka{32$k$}{$d$}
& \blocka{128}{2}\\
  &   & \\
\hline
\multirow{3}{*}{conv4}
& \multirow{3}{*}{8x8} 
& \blocka{64$k$}{$d$}
& \blocka{256}{2}\\
  &   & \\
\hline
& 1x1  & \multicolumn{2}{c}{average pool, 100-$c$ fc, softmax} \\ 
\hline
\end{tabular}
\end{center}
\label{table:WRNBlock}
\vspace{-6mm}
\end{table}

\subsection{Effect of proposed loss components} 
\label{ssec:losses}
Herein, we analyse the overall impact of each constituent loss alongside the optimal approach for combining them. To achieve this, we ran 3 experiments: using $L_{SM}$, using $L_{AdaIN}$ and combining them together: $L_{SM} + L_{AdaIN}$. In order to investigate this in isolation, the losses are always applied on the same location in the network, namely on the output of the conv4 of a WRN for both the student and the teacher (Table~\ref{table:WRNBlock}). 

The results of the above experiments are reported in Table~\ref{table:ablation study}. The results clearly show that all the proposed variants offer significant performance gains: when using $L_{SM}$ and $L_{AdaIN}$ alone, $\sim 1\%$ and $\sim 2\%$ in Top-1 accuracy improvement in absolute terms were obtained. Moreover, when combining the two together, we gained an additional $\sim 0.6\%$ improvement to a total of $2.6\%$. This further shows the importance of transferring statistical information in the form of the teacher's first and second order statistics as proposed by our method, and that the student is able to greatly enhance its accuracy by learning this information from the teacher. Besides, it also suggests that AdaIN in Section \ref{ssec:AIN} is more effective for statistic transfer from the teacher to student. 

\begin{table}[t]
\caption{Effect of proposed losses and position of distillation on the test set of CIFAR-100. $L_{SM}$ and $L_{AdaIN}$ are proposed in this work.}
\vspace{-2mm}
\begin{center}
\begin{tabular}{c|c|c|c}
\hline
Method&Group&Top-1 (\%)&Top-5 (\%)\\
\hline
\multicolumn{2}{c|}{Student (WRN-16-4)} & 75.68&93.15\\
\multicolumn{2}{c|}{Teacher (WRN-40-4)} & 78.31&93.86\\
\hline
$L_{SM}$ &conv4 &76.61&93.73\\
$L_{AdaIN}$&conv4 &77.66&94.20\\
$L_{SM}\mathord{+}L_{AdaIN}$&conv4 &\textbf{78.25}&\textbf{94.59}\\
\hline
$L_{SM}\mathord{+}L_{AdaIN}$&conv2 &75.83&93.28\\
$L_{SM}\mathord{+}L_{AdaIN}$&conv3 &76.28&93.41\\
$L_{SM}\mathord{+}L_{AdaIN}$&conv2+3+4&78.12&94.53\\
\hline
\end{tabular}
\end{center}
\vspace{-6mm}
\label{table:ablation study}
\end{table}

\subsection{Distillation position}
\label{ssec:position}
One crucial aspect, often overlooked in the literature, is the position inside the network where the knowledge distillation losses should be applied. On one hand applying them early in the network (i.e. between corresponding features of the initial layers) could potentially ensure that the subsequent layers in the student network receive ``better'' features. On the other hand, features produced by such early layers typically represent simple geometrical features such as edges that are not specialised to a particular given class of objects. As such, intuitively applying the distillation loss towards the end of the network, where the activations encode high level, discriminative, task-related features should lead to potentially stronger models. 

The results from Table~\ref{table:ablation study} (last 3 rows) confirm our hypothesis: although applying the proposed losses throughout the network always results in better classification performance, the deeper the features used for knowledge distillation are, the higher the obtained gains in accuracy. Ours findings align with those reported in~\cite{yosinski2014transferable}, where Yosinski \etal showed that features from shallow layers are more general while the ones from higher layers have a greater specificity. Also in~\cite{kim2018paraphrasing}, Kim \etal selected features from the last layer. Furthermore, applying the distillation loss at multiple points in the network does not improve the accuracy gains further suggesting that the activations of the last layers alone are sufficiently informative. As such for the remaining of our experiments, the distillation losses were applied to the deeper layers only (i.e. conv4). 

\subsection{Method analysis}
The following two paragraphs present two of the most important experiments presented in our paper. Paragraph \textbf{Teacher-student similarity} shows that the proposed method significantly increases the similarity between the teacher's and student's output. Paragraph \textbf{Statistics Distance} shows that the proposed losses, and especially $L_{AdaIN}$, significantly reduces the distance between the statistics of the teacher and student. Overall, the results of the next paragraphs clearly show that effective statistics transfer, as proposed and described in our method, correlates with training more accurate student models.

\begin{table}[t]
\caption{KL divergence between teacher and student, and cross-entropy between ground truth and student on the test set of CIFAR-100. $L_{SM}$ and $L_{AdaIN}$ are proposed in this work.}
\vspace{-2mm}
\begin{center}
\begin{tabular}{c|c|c|c}
\hline
Method & \tabincell{c}{KL div. with teacher}&\tabincell{c}{Cross-entropy with label}&\tabincell{c}{Top-1 (\%)}\\
\hline
Student &0.6364&0.9981&75.68\\
KD~\cite{hinton2015distilling} & 0.6227 &0.9824&77.17\\
AT~\cite{zagoruyko2016paying} & 0.6122 &0.9741&76.48\\
OFD~\cite{heo2019comprehensive}&0.5348&0.9652&77.79\\
RKD~\cite{park2019relational}&0.5829&0.9583&77.48\\
CRD~\cite{tian2019contrastive}&0.5992&0.9418&77.49\\
\hline
$L_{SM}$ &0.6169&0.9785&76.61\\
$L_{AdaIN}$  &0.5312&0.9122&77.66\\
$L_{SM}\mathord{+}L_{AdaIN}$ &\textbf{0.5160}&\textbf{0.9051}&\textbf{78.25}\\
\hline
\end{tabular}
\end{center}
\label{table:similarity}
\vspace{-4mm}
\end{table}

\noindent\textbf{Teacher-student similarity:} The overall aim of knowledge distillation is to make the student mimic the teacher's output, so that student is able to obtain similar performance as the teacher. Therefore, to see how well the student mimics the teacher, we measure the similarity between the teacher's and student's outputs using (a) the KL divergence between the teacher's and student's outputs, and (b) the cross-entropy loss between the student's predictions and the ground truth label. 

From Table~\ref{table:similarity}, it can be observed that KD~\cite{hinton2015distilling} reduces the KL divergence with the teacher's output offering $\sim 1.5\%$ accuracy gain. AT~\cite{zagoruyko2016paying} (one of the representative methods for feature distillation), also decreases the KL divergence with the teacher's output offering a smaller accuracy gain of $\sim 0.8\%$. Moreover, both proposed losses $L_{SM}$ and $L_{AdaIN}$ and their combination $L_{SM}\mathord{+}L_{AdaIN}$ show considerably high similarity compared to the other methods. This similarity is one of the main reasons for the improved performance achieved by our method. 

\begin{table}[t]
\caption{Distance between the statistics of the teacher and the student calculated on the test set of CIFAR-100. NMI on the test set of CIFAR-100}
\vspace{-2mm}
\begin{center}
\begin{tabular}{c|cccc}
\hline
Method &Student &$L_{SM}$ &$L_{AdaIN}$ &$L_{SM}\mathord{+}L_{AdaIN}$\\
\hline
L2 Distance& 1.44&1.26 &1.02&\textbf{0.94}\\
NMI (\%)&76.10 &77.00 &77.80&\textbf{78.50}\\
Top-1(\%)&75.68&76.61&77.66&\textbf{78.25}\\ 
\hline
\end{tabular}
\end{center}
\label{table:statistic diatance}
\vspace{-4mm}
\end{table}

\begin{figure}[t]
\centering
\subfigure[{\fontsize{6}{7}\selectfont Student}]{\includegraphics[width=0.192\textwidth]{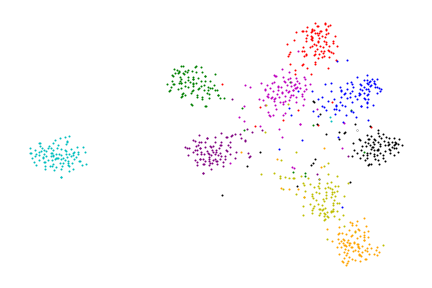}}
\subfigure[{\fontsize{6}{7}\selectfont $L_{SM}$}]{\includegraphics[width=0.192\textwidth]{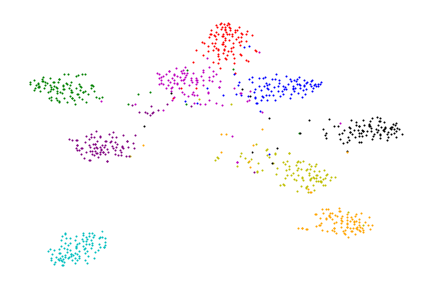}}
\subfigure[{\fontsize{6}{7}\selectfont $L_{AdaIN}$}]{\includegraphics[width=0.192\textwidth]{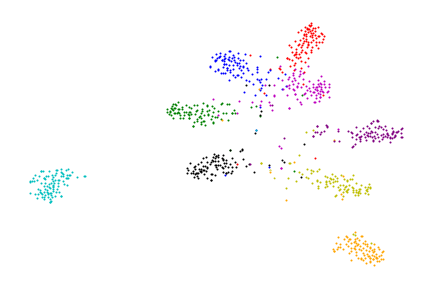}}
\subfigure[{\fontsize{6}{7}\selectfont $L_{SM}\mathord{+}L_{AdaIN}$}]{\includegraphics[width=0.192\textwidth]{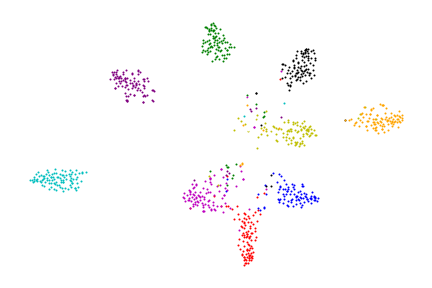}}
\subfigure[{\fontsize{6}{7}\selectfont Teacher}]{\includegraphics[width=0.192\textwidth]{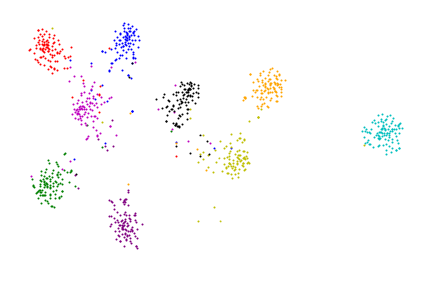}}
\caption{Visualization of feature maps after Conv4 on test set of CIFAR-100. Better viewed in color.}
\label{fig:feature_visualization}
\vspace{-2mm}
\end{figure}

\noindent
\textbf{Statistics Distance:} We calculated the distance between the statistics learned by student network and those of teacher. The results are presented in Table \ref{table:statistic diatance}. It is clear that both proposed losses narrow the statistics distance between the teacher and the student, with their combination $L_{SM}\mathord{+}L_{AdaIN}$ being the closest to the teacher. Therefore, $L_{SM}\mathord{+}L_{AdaIN}$ achieves the best accuracy. Moreover, we calculated the Normalized Mutual Information (NMI)~\cite{manning2008introduction} which is a balanced metric determining the quality of clustering. The results are presented in Table \ref{table:statistic diatance}. The results show that $L_{SM}+L_{AdaIN}$ gets the highest NMI score which means that the features are better clustered. Qualitative results are shown in Figure~\ref{fig:feature_visualization} which visualizes the feature maps after conv4 layer in both the student and the teacher. It can be observed that $L_{SM}\mathord{+}L_{AdaIN}$ is able to learn more discriminative features and achieves the best discrimination which also correlates with quantitative performance.

\section{Comparison with state-of-the-art}
In this section, we thoroughly evaluate the effectiveness of the proposed method across multiple (a) network architectures (ResNet~\cite{He_2016_CVPR}, Wide ResNet~\cite{zagoruyko2016wide}, MobileNetV2~\cite{mobilenetv2}, MobileNet~\cite{howard2017mobilenets}) with different teacher-student capacities, (b) datasets (CIFAR10/100, ImageNet), and (c) domains (real-valued and binary networks). 
For the above mentioned settings, we compare our method with KD~\cite{hinton2015distilling} and AT~\cite{zagoruyko2016paying}, and the more recent methods of OFD~\cite{heo2019comprehensive} from ICCV 2019, RKD~\cite{park2019relational} from CVPR2019, CRD~\cite{tian2019contrastive} from ICLR2020.  

\noindent\textbf{Overview of results:} From our experiments below, we can draw the following conclusion: our approach 
offers consistent gains across all of the above scenarios, outperforming all methods considered for all settings. Notably, our method is particularly effective for the most difficult datasets (i.e. CIFAR-100 and ImageNet). 
 
\noindent \textbf{Network architectures:} Following prior work (see Section~\ref{section:Related Work}), we mainly conducted experiments using variants of the ResNet~\cite{He_2016_CVPR} and WRN~\cite{zagoruyko2016wide} architectures. Throughout this work, following the convention proposed in~\cite{He_2016_CVPR}, we denote with ResNet-N a Residual Network with N convolutional layers. WRNs expand upon the original ResNet by proposing to increase the overall width by $k$ times. Following~\cite{zagoruyko2016paying}, we denote with WRN-$D$-$k$ a WRN architecture with $D$ layers and an expansion rate of $k$. 

\subsection{CIFAR-10}
\noindent\textbf{Implementation details:} 
CIFAR-10 is a popular image classification dataset consisting of 50,000 training and 10,000 testing images equally distributed across 10 classes. All images are at a resolution of $32\times 32$px. Following~\cite{zagoruyko2016paying}, during training, we randomly cropped and horizontally flipped the images. The ResNet models were trained for 350 epochs using SGD. The initial learning rate was set to 0.1, that is then reduced by a factor of 10 at epochs 150, 250 and 320.
Similarly, the WRN models were trained for 200 epochs with a learning rate of 0.1 that is subsequently reduced by 5 at epochs 60, 120 and 160. In all experiments, we set the dropout rate to 0.
For traditional KD~\cite{hinton2015distilling}, we set $\alpha=0.9$ and $T=4$. For AT~\cite{zagoruyko2016paying}, as in~\cite{zagoruyko2016paying,tung2019similarity} we set the weight of distillation loss to 1000. 
We note, that for our experiments, the AT loss is added after each layer group for WRN and the last two groups for ResNet as in~\cite{zagoruyko2016paying}.
Following OFD~\cite{heo2019comprehensive}, we set the weight of distillation loss to $10^{-3}$.
For RKD~\cite{park2019relational}, we set $\beta_{1} = 25$ for distance, and $\beta_{2} = 50$ for angle, as described in ~\cite{park2019relational,tian2019contrastive}.
We did not compare with CRD~\cite{tian2019contrastive} here because, in our experiments, we found that their parameter setting used for CIFAR-100 and ImageNet-1K in their paper struggles to obtain good performance in CIFAR-10.

\noindent\textbf{Results:} Our results are reported in Table \ref{table:result_cifar10}. Note that, for CIFAR-10, only Top-1 accuracy is reported (Top-5 does not make sense for 10 classes). We tested three cases representing different network architectures for student and teacher networks: the first two experiments are with WRNs. The following three experiments are with ResNets. In the last experiment, teacher and student have different network architectures. Overall, our method achieves the best results for all cases, with KD~\cite{hinton2015distilling} closely following. In all following experiments though, the gap between our method and KD becomes significantly larger. 

\begin{table}[t]
\caption{Top-1 accuracy ($\%$) of various knowledge distillation methods on CIFAR-10.}
\vspace{-2mm}
\begin{center}
\resizebox{1\textwidth}{!}{
\setlength\tabcolsep{3.5pt}
\begin{tabular}{ll|cccccc|c}
\hline
Student(Params) & Teacher(Params) & Student & KD~\cite{hinton2015distilling} & AT~\cite{zagoruyko2016paying} &OFD~\cite{heo2019comprehensive}&RKD~\cite{park2019relational}&Ours & Teacher\\
\hline
WRN-16-1 (0.18M) & WRN-16-2 (0.69M)& 91.04 & 92.57 & 92.15 &92.28&92.51&\textbf{92.90}&93.98\\
WRN-16-2 (0.69M) & WRN-40-2 (2.2M)&93.98&94.46&94.39 &94.30&94.41&\textbf{94.67}&95.07\\
\hline
ResNet-8\enspace (0.08M) & ResNet-26 (0.37M) & 87.78 & 88.75 & 88.15 &87.49&88.50& \textbf{89.02} & 93.58\\
ResNet-14 (0.17M) & ResNet-26 (0.37M) & 91.59 & 92.57 & 92.11 
&92.51&92.36&\textbf{92.80} & 93.58\\
ResNet-18 (0.7M) & ResNet-34 (1.4M) & 93.35 & 93.74 & 93.52 
&93.80&92.95& \textbf{93.92} & 94.11\\
\hline
WRN-16-1 (0.18M) & ResNet-26 (0.37M)&91.04&92.42&91.32&92.47&92.08&\textbf{92.94}&93.58\\
\hline
\end{tabular}}
\end{center}
\label{table:result_cifar10}
\vspace{-8mm}
\end{table}

\subsection{CIFAR-100} 
\label{ssec:CIFAR100}
\noindent\textbf{Implementation details:} CIFAR-100~\cite{krizhevsky2009learning} is similar to CIFAR-10 but has 100 classes, containing only 500 images per classes. We used a standard data augmentation scheme~\cite{zagoruyko2016paying} including padding 4 pixels before random cropping, horizontal flipping, and 15 degree rotation. We used SGD with weight decay 5e-4 and momentum 0.9 as optimizer. Batch size was set 128. The other experimental settings are as in CIFAR-10.\\  
\noindent\textbf{Results:} For CIFAR-100, Top-1 performance of our method, as well as for different teacher-student methodologies, is presented in Table~\ref{table:result_cifar100}. 
Top-5 is provided in the supplementary material. 
We experimented with several student-teacher network pairs using different structures. Experiments are grouped in three sets. The first shows performance for different teacher and student capacities using WRNs: poor student - good teacher (WRN-16-2; WRN-40-4), descent student - good teacher (WRN-10-10; WRN-16-10); good student - good teacher (WRN-16-4; WRN-40-4). In a similar fashion, in the second set, we show that these results hold when using a different architecture, ResNet in this case. The final set is designed to show the performance when teacher and student have different architectures. We used MobileNetV2, ResNet and WRN to this end.

We observe that for almost all configurations, our method achieves consistent and significant accuracy improvements over prior work, overall being the best performing method.
Furthermore, it is hard to tell which is the second best method as the remaining methods have their own advantages for different configurations. For WRN experiments, OFD ranks second. For ResNet experiments, CRD ranks second. 
For mixed structures, CRD and OFD perform similarly. We also report more comparisons with other methods and the results obtained by combining our method with KD and AT in the supplementary material.
We believe that further improvements could be obtained by combining our method with other methods but this requires a comprehensive investigation which goes beyond the scope of this paper. 

\begin{table}[t]
\caption{Top-1 accuracy ($\%$) of various knowledge distillation methods on CIFAR-100.} 
\vspace{-3mm}
\begin{center}
\resizebox{1\textwidth}{!}{
\setlength\tabcolsep{3.5pt}
\begin{tabular}{ll|ccccccc|c}
\hline
Student (Params) & Teacher (Params) & Student & KD~\cite{hinton2015distilling} &AT~\cite{zagoruyko2016paying}
&OFD~\cite{heo2019comprehensive}&RKD~\cite{park2019relational}&CRD~\cite{tian2019contrastive}&Ours& Teacher \\
\hline
WRN-16-2 (0.70M) & WRN-40-4 (8.97M) & 72.22 & 73.32 & 72.56 
&74.12&73.08&73.75&\textbf{74.58} & 78.31\\

WRN-16-4 (2.77M) & WRN-40-4 (8.97M) & 75.68 & 77.17 & 76.48 
&77.79&77.48&77.49&\textbf{78.25}& 78.31\\

WRN-10-10 (7.49M) & WRN-16-10 (17.2M) & 73.75 & 76.63 & 74.63
&76.97&76.28&76.28& \textbf{77.12}  & 78.75\\

\hline
ResNet-10 (0.34M) & ResNet-34 (1.39M) & 67.60 & 68.05 & 68.39
&67.94&68.43&69.17&\textbf{69.49} & 72.05\\

ResNet-18 (0.75M) & ResNet-50 (1.99M) & 70.21 & 72.40 & 70.78
&71.54&72.10&72.31&\textbf{72.59}& 72.83\\

ResNet-10 (4.95M) & ResNet-34 (21.33M) & 74.20 & 75.79 & 75.03
&75.60&75.04&\textbf{76.63}& 76.11  & 77.26\\

\hline
WRN-16-2 (0.70M) & ResNet-34 (21.33M) & 72.22 & 73.32 & 71.46 
&73.78&72.95&73.39& \textbf{74.38} & 77.26\\
MobileNetV2 (2.37M)&ResNet-34 (21.33M)&68.36&68.65&66.68
&70.18&69.52&\textbf{70.98}&70.66&77.26\\
MobileNetV2 (2.37M)&WRN-40-4 (8.97M) & 68.36 & 68.40&66.46
&70.10&68.64&70.45&\textbf{71.15}&78.31\\
\hline
\end{tabular}}
\end{center}
\label{table:result_cifar100}
\vspace{-4mm}
\end{table}

\subsection{ImageNet-1K}
\noindent
\textbf{Implementation Details:} ImageNet-1K~\cite{ILSVRC15} is a large classification dataset which consists of 1.2M training images and 50K validation images with 1,000 classes. Images are cropped to $224\times 224$ pixels for both training and evaluation.
We used SGD with Nesterov momentum 0.9, initial learning rate 0.2, weight decay $1e-4$, and dropped the learning rate by 0.1 every 30 epochs, training in total for 100 epochs except CRD with 10 more epochs as is suggested by its authors. Batch size was set to 512. For simplicity and to enable a fair comparison, we used pretrained PyTorch models~\cite{paszke2017automatic} as teacher networks as adopted in ~\cite{heo2019comprehensive,tian2019contrastive}. Our experiments include two pairs of networks which are popular settings for ImageNet-1K. The first experiment is distillation from ResNet-34 to ResNet-18 and the second one is distillation from ResNet-50 to MobileNet~\cite{howard2017mobilenets}. Note that following~\cite{tian2019contrastive} on ImageNet, for KD, we set the weight for KL loss as 0.9, the weight for cross-entropy loss as 0.5 which helps to obtain better accuracy. 

\noindent
\textbf{Results:} The comparisons are presented in Table \ref{table:imagenet}. Again, we observe that our method achieves significant improvement over all methods. Moreover, there is no method which is consistently second: for ResNet-34 to ResNet-18 experiment, RKD is the second best while for ResNet-50 to MobileNet, CRD is the second best. Notably, for the latter experiment, CRD method reduces the gap between the teacher and the student by $1.27\%$, while our method narrows it by $2.36\%$. Overall, our results on ImageNet validate the scalability of our method.  

\begin{table}[t]
\caption{Comparison with state-of-the-art on ImageNet.}
\vspace{-4mm}
\begin{center}
\resizebox{1\textwidth}{!}{
\setlength\tabcolsep{3.5pt}
\begin{tabular}{ll|c|ccccccc|c}
\hline
Student (Params)    & Teacher (Params)  && Student&KD~\cite{hinton2015distilling} & AT~\cite{zagoruyko2016paying}&OFD~\cite{heo2019comprehensive}
&RKD~\cite{park2019relational} &CRD~\cite{tian2019contrastive}& Ours & Teacher\\ \hline
ResNet18 (11.69M)   & ResNet34 (21.80M) & \begin{tabular}[c]{@{}l@{}}Top-1\\ Top-5\end{tabular} & \begin{tabular}[c]{@{}l@{}}70.04\\ 89.48\end{tabular} & \begin{tabular}[c]{@{}l@{}}70.68\\ 90.16\end{tabular} & \begin{tabular}[c]{@{}l@{}}70.59\\ 89.73\end{tabular} & \begin{tabular}[c]{@{}l@{}}71.08\\ 90.07\end{tabular}   & \begin{tabular}[c]{@{}l@{}}71.34\\ 90.37\end{tabular} & \begin{tabular}[c]{@{}l@{}}71.17\\ 90.13\end{tabular} & \begin{tabular}[c]{@{}l@{}}\textbf{71.82}\\ \textbf{90.58}\end{tabular} & \begin{tabular}[c]{@{}l@{}}73.31\\ 91.42\end{tabular} \\ \hline
MobileNet (4.23M) ) & ResNet50 (25.56M  & \begin{tabular}[c]{@{}l@{}}Top-1\\ Top-5\end{tabular} & \begin{tabular}[c]{@{}l@{}}70.13\\ 89.49\end{tabular} & \begin{tabular}[c]{@{}l@{}}70.68\\ 90.30\end{tabular} & \begin{tabular}[c]{@{}l@{}}70.72\\ 90.03\end{tabular} & \begin{tabular}[c]{@{}l@{}}71.25\\ 90.34\end{tabular} & \begin{tabular}[c]{@{}l@{}}71.32\\ 90.62\end{tabular}& \begin{tabular}[c]{@{}l@{}}71.40\\ 90.42\end{tabular}& \begin{tabular}[c]{@{}l@{}}\textbf{72.49}\\ \textbf{90.92}\end{tabular} & \begin{tabular}[c]{@{}l@{}}76.16\\ 92.86\end{tabular} \\ \hline
\end{tabular}}
\label{table:imagenet}
\end{center}
\vspace{-4mm}
\end{table}
\vspace{-4mm}

\subsection{Binarization Experiments}
\noindent
\textbf{Implementation details:} Training highly accurate binary neural networks is a very challenging task~\cite{rastegari2016xnor,bulat2019xnor}, and to this end, knowledge distillation appears to be a promising direction. In this section, we present results by applying distillation for the task of training binary student networks guided by real-valued teacher networks. The network architecture is kept the same for both the student and the teacher in this case, specifically we used a ResNet using the modifications described in~\cite{bulat2019xnor} for both the binary student and the real-valued teacher. For training, we used Adam as the optimizer with initial learning 0.001 which is reduced by a factor 10 at epochs 150, 250, and 320, training in total for 350 epochs on CIFAR-100 and with initial learning 0.002 which is reduced by a factor at epochs 30, 60, and 90, training in total for 100 epochs on ImageNet-1K. 

Note that, it was not clear to us where to place the distillation position for OFD, so although we included our result for this method we emphasize that this result might be suboptimal.
\\
\noindent
\textbf{Results:} Table~\ref{table:binaty} presents our results. Again, we observe that our method outperforms all methods considerably, showing that it can effectively transfer knowledge between different domains. 
\begin{table}[t]
\caption{Real-to-binary distillation results on CIFAR-100: a real-valued ResNet-34 teacher is used to distill a binary student ResNet-34. Real-to-binary distillation results on ImageNet-1K: a real-valued ResNet-18 is used to distill a binary student. OFD result might be suboptimal.}
\vspace{-2mm}
\begin{center}
\resizebox{1\textwidth}{!}{
\setlength\tabcolsep{3.5pt}
\begin{tabular}{c|c|ccccccc|c}
\hline
Dataset&Method&Binary&KD~\cite{hinton2015distilling} & AT~\cite{zagoruyko2016paying}&OFD~\cite{heo2019comprehensive}&RKD~\cite{park2019relational}&CRD~\cite{tian2019contrastive} &Ours&Real\\
\hline
CIFAR-100&ResNet34&65.34&68.64&68.54&66.84&68.47&67.76&\textbf{70.15}&75.08\\
ImageNet-1K&ResNet18&56.70&57.39&58.45&55.74&58.84&58.25&\textbf{59.87}&70.20\\
\hline
\end{tabular}}
\end{center}
\vspace{-5mm}
\label{table:binaty}
\end{table}

\section{Conclusion}
We presented a method for knowledge distillation based on transferring feature statistics from the teacher to the student. To this end, and going beyond using an $L_2$ loss, the learned statistics, namely the channel-wise mean and variance, are transferred back to the teacher via adaptive instance normalization and evaluated via a loss which enforces that they are reliably transferred. We show that our method, enabled by our newly proposed loss, consistently outperforms other state-of-the-art distillation methods for a wide range of experimental settings including multiple network architectures (ResNet, Wide ResNet, MobileNet) with different teacher-student capacities, datasets (CIFAR10/100, ImageNet), and domains (real-valued and binary networks).  

{\small
\bibliographystyle{splncs}
\bibliography{egbib}
}
\end{document}